# Estimating Motion with Principal Component Regression Strategies

Felipe P. do Carmo [1], Vania Vieira Estrela [*2], Joaquim Teixeira de Assis [3]

*State University of Rio de Janeiro (UERJ), Polytechnic Institute of Rio de Janeiro (IPRJ), CP 972825, CEP 28601-970, Nova Friburgo, RJ, Brazil*

[1] fpcarmo@iprj.uerj.br
[2] vestrela@iprj.uerj.br
[3] joaquim@iprj.uerj.br

*Abstract*— In this paper, two simple principal component regression methods for estimating the optical flow between frames of video sequences according to a pel-recursive manner are introduced. These are easy alternatives to dealing with mixtures of motion vectors in addition to the lack of prior information on spatial-temporal statistics (although they are supposed to be normal in a local sense). The 2D motion vector estimation approaches take into consideration simple image properties and are used to harmonize regularized least square estimates. Their main advantage is that no knowledge of the noise distribution is necessary, although there is a underlying assumption of localized smoothness. Preliminary experiments indicate that this approach provides robust estimates of the optical flow.

## I. INTRODUCTION

In video sequences, motion provides important information. Significant events, such as collision paths, object docking, sensor obstruction, object properties and occlusion can be characterized and better understood with the help of optic flow (OF). Segmenting an OF field (OFF) into coherent motion groups and estimating each underlying motion are very challenging tasks when a scene has several independently moving objects. The problem is further complicated by data that are noisy, and/or partially incorrect (incomplete). Regression models can help suppressing some gross data errors or outliers. However, segmenting an OFF consisting of a large portion of incorrect data or multiple motion groups requires high sturdiness that is unattainable by conventional robust estimators.

The main problem of motion analysis is the difficulty of getting accurate motion estimates without prior motion segmentation and vice-versa. Some researchers tried to address both problems resourcing to unified frameworks which allow simultaneous motion estimation and segmentation. One method uses a robust estimator which can cope with errors due to noise and scene clutter (multiple moving objects). It is based on the Hough transform implemented in a search mode. This has the benefit of identifying the most significant moving regions first and, thus, providing an effective focus of attention mechanism. The motion estimator and segmentor also provides information about the confidence of motion estimates. This procedure is very useful not only from the point of view of scene interpretation, but also from the point of other applications such as video compression and coding.

In coding applications, a block-based approach is often used for interpolation of lost information between key frames [14]. The fixed rectangular partitioning of the image used by some block-based approaches often separates visually meaningful image features. Pel-recursive (PR) schemes ([6, 7, 13, 14]) can theoretically overcome some of the limitations associated with blocks by assigning a unique motion vector to each pixel. Intermediate frames are then constructed by resampling the image at locations determined by linear interpolation of the motion vectors. The PR approach can also manage motion with sub-pixel accuracy. The update of the motion estimate was based on the minimization of the displaced frame difference (DFD) at a pixel. In the absence of additional assumptions about the pixel motion, this estimation problem becomes "ill-posed" because of the following problems: a) occlusion; b) the solution to the 2D motion estimation problem is not unique (aperture problem); and c) the solution does not continuously depend on the data due to the fact that motion estimation is highly sensitive to the presence of observation noise in video images.

Segmenting OF via expectation maximization (EM) for mixtures of principal component analysis (PCA) because both techniques share a close relationship can be done successfully [4, 15]. An approach called generalized PCA (GPCA) models the OF from scenes containing dynamic textures [18] and it does not require any initialization. This approach first projects the data points onto a low-dimensional subspace, then a polynomial is fit to the projected data points and a basis for each one of the projected subspaces is obtained from the derivatives of this polynomials at the data points.

Most methods assume that there is little or no interference between the individual sample constituents or that all the constituents in the samples are known ahead of time. In real world samples, it is very unusual, if not entirely impossible, to know the entire composition of a mixture sample. Sometimes, only the quantities of a few constituents in very complex mixtures of multiple constituents are of interest ([2, 13, 15]).

This work intends to solve OF problems by means of two different takes on PCA regression (PCR): 1) a combination of regularized least squares (RLS) and PCA ($PCR_1$); and 2) RLS followed by regularized PCA regression ($PCR_2$). Both involve simpler computational procedures than previous attempts at addressing mixtures [2, 12, 13, 15, 16].

Section II sets up a model for the OF estimation problem and it states two forms of regression for the computation of motion: the ordinary least squares (OLS); and, one of its extensions - the regularized least squares - RLS, ([5, 8-12, 14]). Section III comments PCA in the context of OF, so that the proposed techniques ($PCR_1$ and $PCR_2$) can be formulated. Section IV shows some experiments used to access their performance. Finally, a discussion of the results and future research plans are presented in Section V.

## II. PROBLEM FORMULATION

The displacement of every pixel in each frame forms the displacement vector field (DVF) and its estimation can be done using at least two successive frames. A vector is assigned to each point in the image when a pixel belongs to a moving area, if its intensity has changed between consecutive frames. Hence, our goal is to find the corresponding intensity value $I_k(\mathbf{r})$ of the *k*-th frame at location $\mathbf{r} = [x, y]^T$, and $\mathbf{d}(\mathbf{r}) = [d_x, d_y]^T$ the corresponding displacement vector (DV) at the working point **r** in the current frame through PR algorithms.

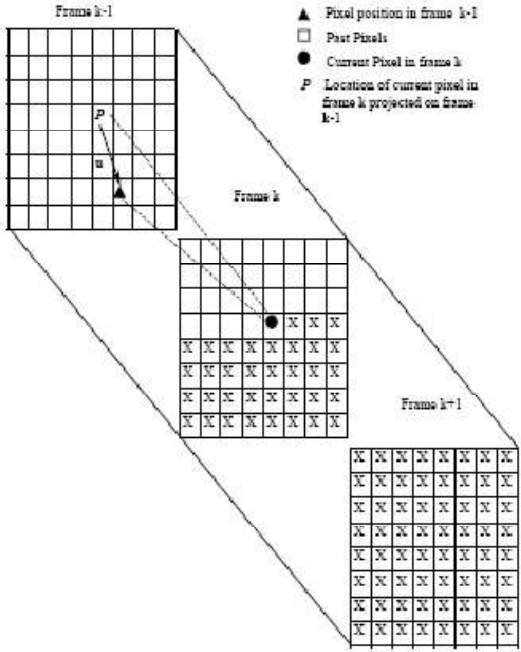

Fig. 1. Backward motion estimation problem.

PR algorithms minimize the DFD function in a small area containing the working point assuming constant image intensity along the motion trajectory. The perfect registration of frames will result in $I_k(\mathbf{r})=I_{k-1}(\mathbf{r}-\mathbf{d}(\mathbf{r}))$. Figure 1 shows some examples of pixel neighbourhoods. The DFD represents the error due to the nonlinear temporal prediction of the intensity field through the DV and is given by

$$\Delta(\mathbf{r};\mathbf{d}(\mathbf{r}))=I_k(\mathbf{r})-I_{k-1}(\mathbf{r}-\mathbf{d}(\mathbf{r})) \quad . \quad (1)$$

In this text, OF is the 2D field of instant velocities or, equivalently, displacement vectors (DVs), of brightness patterns in the image plane. PR algorithms are recursive predictor-corrector-type of estimators [14]. They start from an initial estimation made for a given point by prediction from a previous pixel or set of pixels or from some other prediction scheme. Later, this estimate is corrected according to some metrics and/or other criteria.

The relationship between the DVF and the intensity field is nonlinear. An estimate of $\mathbf{d}(\mathbf{r})$, is obtained by directly minimizing $\Delta(\mathbf{r},\mathbf{d}(\mathbf{r}))$ or by determining a linear relationship between these two variables through some model. This is accomplished by using a Taylor series expansion of $I_{k-1}(\mathbf{r}-\mathbf{d}(\mathbf{r}))$ about the location $(\mathbf{r}-\mathbf{d}^i(\mathbf{r}))$, where $\mathbf{d}^i(\mathbf{r})$ represents a prediction of $\mathbf{d}(\mathbf{r})$ in i-th step. This results in $\Delta(\mathbf{r},\mathbf{r}-\mathbf{d}^i(\mathbf{r})) = -\mathbf{u}^T \nabla I_{K-1}(\mathbf{r}-\mathbf{d}^i(\mathbf{r}))+e(\mathbf{r},\mathbf{d}(\mathbf{r}))$, where the displacement update vector is $\mathbf{u}=[u_x, u_y]^T = \mathbf{d}(\mathbf{r}) - \mathbf{d}^i(\mathbf{r})$, $e(\mathbf{r}, \mathbf{d}(\mathbf{r}))$ stands for the truncation error resulting from higher order terms (linearization error) and $\nabla=[\partial/\partial_x, \partial/\partial_y]^T$ represents the spatial gradient operator. Applying Eq.(1) to all points in a neighbourhood of pixels $\mathcal{R}$ around **r** gives

$$\mathbf{z} = \mathbf{Gu} + \mathbf{n} \quad , \quad (2)$$

where the temporal gradients $\Delta(\mathbf{r}, \mathbf{r}-\mathbf{d}^i(\mathbf{r}))$ have been stacked to form the $N \times 1$ observation vector **z** containing DFD information on all the pixels in $\mathcal{R}$, the $N \times 2$ matrix **G** is obtained by stacking the spatial gradient operators at each observation, and the error terms have formed the $N \times 1$ noise vector **n**. The PR estimator for each pixel located at position **r** of a frame *k* can be written as

$$\mathbf{d}^{i+1}(\mathbf{r}) = \mathbf{d}^i(\mathbf{r}) + \mathbf{u}^i(\mathbf{r}),$$

where $\mathbf{u}^i(\mathbf{r})$ is the current motion update vector obtained through a motion estimation procedure that attempts to solve (2), $\mathbf{d}^i(\mathbf{r})$ is the DV at iteration *i* and $\mathbf{d}^{i+1}(\mathbf{r})$ is the corrected DV.

This work concentrates its attention on regression-like methods to solve Eq. (2) for **u**. These algorithms use the matrix $p \times p$ $\mathbf{G}^T\mathbf{G}$ in one way or another.

The ordinary least squares (OLS) estimate of the update vector is

$$\mathbf{u}_{LS} = (\mathbf{G}^T\mathbf{G})^{-1}\mathbf{G}^T\mathbf{z} \quad ,$$

which is given by the minimizer of the functional $J(\mathbf{u})=\|\mathbf{z}-\mathbf{Gu}\|^2$ (for more details, see [4, 5, 7]). The assumptions made about **n** for least squares estimation are $E(\mathbf{n}) = \mathbf{0}$, and Var(**n**)

= $E(\mathbf{n}\mathbf{n}^T) = \sigma^2 \mathbf{I}_N$, where $E(\mathbf{n})$ is the expected value (mean) of **n**, and $\mathbf{I}_N$, is the identity matrix of order *N*. From now on, **G** will be analyzed as being an $N \times p$ matrix in order to make the whole theoretical discussion easier. Since **G** may be very often ill–conditioned, the solution given by the previous expression will be usually unacceptable due to the noise amplification resulting from the calculation of the inverse matrix $\mathbf{G}^T\mathbf{G}$. In other words, the data are erroneous or noisy. Hence, one cannot expect an exact solution for **z = Gu + n**, but rather an approximation according to some course of action.

The regularized minimum norm solution to Eq. (2) - also known as regularized least square (RLS) solution – is given by

$$\hat{\mathbf{u}}_{RLS}(\Lambda) = (\mathbf{G}^T\mathbf{G} + \Lambda)^{-1}\mathbf{G}^T\mathbf{z}.$$

In order to improve the RLS estimate of the motion update vector, we propose a strategy which takes into consideration the local properties of the image. It is described in the next section.

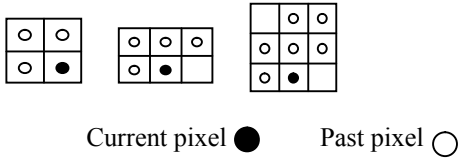

Current pixel ●     Past pixel ○

Figure 2. Examples of causal masks.

Each row of **G** has entries $[g_{xi}, g_{yi}]^T$, with *i = 1, …, N*. The spatial gradients of $I_{k-1}$ are calculated through a bilinear interpolation scheme similar to what is done in [4, 5].

The entries $f_{k-1}(\mathbf{r})$ corresponding to a given pixel location inside a causal mask are needed to compute the spatial gradients by means of bilinear interpolation [4, 5] at location $\mathbf{r} = [x, y]^T$ as follows:

$$\begin{bmatrix} \theta_x \\ \theta_y \end{bmatrix} = \begin{bmatrix} x - \lfloor x \rfloor \\ y - \lfloor y \rfloor \end{bmatrix},$$

where $\lfloor x \rfloor$ is the largest integer that is smaller than or equal to *x*, the bilinear interpolated intensity $f_{k-1}(\mathbf{r})$ is specified by

$$f_{k-1}(\mathbf{r}) = \begin{bmatrix} 1 - \theta_x \\ \theta_x \end{bmatrix}^T \begin{bmatrix} f_{00} & f_{10} \\ f_{01} & f_{11} \end{bmatrix} \begin{bmatrix} 1 - \theta_y \\ \theta_y \end{bmatrix},$$

with $f_{ij}(\mathbf{r}) = f_{k-1}(\lfloor x \rfloor + i, \lfloor y \rfloor + j)$. The equation above can be used for evaluating the second order spatial derivatives of $f_{k-1}(\mathbf{r})$ at **r** by means of backward differences:

$$\begin{bmatrix} g_x \\ g_y \end{bmatrix} = \begin{bmatrix} (1-\theta_y)(f_{01} - f_{00}) + (f_{11} - f_{10})\theta_y \\ (1-\theta_x)(f_{10} - f_{00}) + (f_{11} - f_{01})\theta_x \end{bmatrix}.$$

### III. ON THE USE OF PRINCIPAL COMPONENT ANALYSIS IN REGRESSION

The main idea behind the two proposed PCR procedures is the PCA of the **G** matrix [10-12]. They yield the same result, but differ in accuracy, and computing time.

PCA is a useful method to solve problems including exploratory data analysis, classification, variable decorrelation prior to the use of neural networks, pattern recognition, data compression, and noise reduction, for example. The formulation of PCA implies a Gaussian latent variable model and can easily lead to Bayesian models.

This technique is used whenever uncorrelated linear combinations of variables are wanted which reduces the dimensions of a set of variables by reconstructing them into uncorrelated combinations. It combines the variables that account for the largest part of the variance to form the first PC. The second PC accounts for the next largest amount of variance, and so on, until the complete sample set variance is combined into progressively smaller uncorrelated component categories. Each successive component explains portions of the variance in the total sample. PCA relates to the second statistical moment of **G**, which is proportional to $\mathbf{G}^T\mathbf{G}$ and it partitions **G** into matrices **T** and **P** (sometimes called scores and loadings, respectively), such that:

$$\mathbf{G} = \mathbf{TP}^T.$$

Matrix **T** contains the eigenvectors of $\mathbf{G}^T\mathbf{G}$ ordered by their eigenvalues with the largest first and in descending order. If **P** has the same rank as **G**, i.e., **P** contains the eigenvectors to all nonzero eigenvalues, then **T = GP** is a rotation of **G**. The first column of **P**, $\mathbf{p}_1$, gives the direction that minimizes the orthogonal distances from the samples to their projection onto this vector. This means that the first column of *T* represents the largest possible sum of squares as compared to any other direction in $\mathbb{R}^N$. It is customary to center the variables in matrix *G* prior to using PCA. This makes $\mathbf{G}^T\mathbf{G}$ proportional to the variance-covariance matrix. The first principal axis is then the direction in which the data have the largest spread. *T* and *P* can be found by means of singular value decomposition. When dimensionality reduction is needed, the number of components can be chosen via examination of the eigenvalues or, for instance, considering the residual error from cross-validation ([5, 12]). Due to the nature of our stated motion estimation problem, the PCs will be kept and used to group displacement vectors inside a neighbourhood. The resulting clusters give an idea about the mixture of motion vectors inside a mask.

As its name implies, this method is closely related to principal components analysis. In essence, it is just multiple linear regression of PCA scores on **z**. The formal solution may be written as [19]:

$$\hat{\mathbf{u}}_{PCR1} = \mathbf{P}(\mathbf{T}^T\mathbf{T})^{-1}\mathbf{T}^T\mathbf{z}.$$

The previous expression is called PCR$_1$. It should be pointed out that the inverse is stabilized in an altogether different way from philosophy behind the regularized least squares (RLS) solution

$$\hat{\mathbf{u}}_{RLS}(\mathbf{\Lambda}) = (\mathbf{G}^T\mathbf{G} + \mathbf{\Lambda})^{-1}\mathbf{G}^T\mathbf{z},$$

where a regularization matrix $\mathbf{\Lambda}$ tries to compensate for deviations from the smoothness constraint.

Returning to PCR$_1$, the scores vectors (columns in $\mathbf{T}$) of different components are orthogonal. PCR$_1$ uses a truncated inverse where only the scores corresponding to large eigenvalues are included. The main drawback of PCR$_1$ is that the largest variation in $\mathbf{G}$ might not correlate with $\mathbf{z}$ and therefore the method may require the use of a more complex model. Some nice properties of the PCA are:

1) If the complete set of PCs is used, PCA will produce the same results as the original OLS, but with possibly more accuracy, if the original $\mathbf{G}^T\mathbf{G}$ matrix has inversion problems.

2) If $\mathbf{G}^T\mathbf{G}$ is nearly singular, a solution better than the one given by the OLS can be obtained by means of a reduced set of PCs due to the calculated variances.

3) Since the PCs are uncorrelated, straightforward significance tests may be employed that do not need be concerned with the order in which the PCs were entered into the regression model. The regression coefficients will be uncorrelated and the amounts explained by each PC are independent and hence additive so that the results may be reported in the form of an analysis of variance.

4) If the PCs can be easily interpreted, the resultant regression equations may be more meaningful.

The criteria for deciding when the PCR$_1$ estimator is superior to OLS estimators depend on the values of the true regression coefficients in the model.

The previous solution can also be regularized:

$$\hat{\mathbf{u}}_{PCR2} = \mathbf{P}(\mathbf{T}^T\mathbf{T} + \mathbf{\Xi})^{-1}\mathbf{T}^T\mathbf{z},$$

with $\mathbf{\Xi}$ standing for a regularization matrix in the PC domain.

## IV. EXPERIMENTS

Grouping objects can be posed as a mathematical problem consisting of finding region boundaries. This discrimination analysis (DA) among objects can be highly nonlinear.

Sometimes the problem is such that a sample may belong to more than one class at the same time, or not belong to any class. In this method each class is modelled by a multivariate normal in the score space from PCA. Two measures are used to determine whether a sample belongs to a specific class or not. One is the leverage or the Mahalanobis distance to the center of the class, the class boundary being computable as an ellipse (please, see Fig. 3). The other is the norm of the residual, which must be lower than a critical value.

In Fig.3, a set of observations is plotted with respect to the first two principal components (PCs). One can easily apprehend that there is a strong suggestion of four distinct groups on which convex hulls and ellipses have been drawn around the four suspected groups. It is likely that the four clusters shown correspond to four different types of displacement vectors. For a big neighbourhood, it could happen that these vectors would not be readily distinguished using only one variable at a time, but the plot with respect to the two PCs clearly distinguishes the four populations.

PCA provides additional information about the data being analyzed. The eigenvalues of the correlation matrix of predictor variables play an important role in detecting multi-collinearity and in analyzing its effects. The PCR estimates are biased, but may be more accurate than OLS estimates in terms of mean square error. It is only possible to evaluate the gain in accuracy for the two new methods, compared to OLS and RLS, for synthetic video sequences, since knowledge of the true values of the coefficients is required. Nevertheless, when severe multi-collinearity is suspected, it is recommended that at least one set of estimates in addition to the OLS estimates be computed since these estimates may help interpreting the data in a different way.

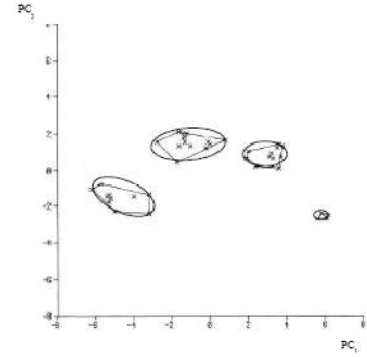

Fig. 3. An example of cluster analysis obtained by means of principal components.

When PCA reveals the instability of a particular data set, one should first consider using least squares regression on a reduced set of variables. If least squares regression is still unsatisfactory, only then should principal components be used. Besides exploring the most obvious approach, it reduces the computer load. Outliers and other observations should not be automatically removed, because they are not necessarily bad observations. As a matter of fact, they can signal some change in the scene context and if they make sense according to the above-mentioned criteria, they may be the most informative points in the data. For example, they may indicate that the data did not come from a normal population or that the model is not linear.

When cluster analysis is used for video scene dissection, the aim of a two-dimensional plot with respect to the first two PCs will almost always be to verify that a given dissection 'looks' reasonable. Hence, the diagnosis of areas containing motion discontinuities can be significantly improved. If additional knowledge on the existence of borders is used, then one's ability to predict the correct motion will increase.

PCs can be used for clustering, given the links between regression and discrimination. The fact that separation among populations may be in the directions of the last few PCs does not mean that PCs should not be used at all.

In regression, their uncorrelatedness implies that each PC can be assessed independently. This is an advantage compared to using the original variables, where the contribution of one of the variables depends on which other variables are also included in the analysis, unless all elements are uncorrelated. To classify a new observation, the least distance cluster is picked up. If a datum is not close to any of the existing groups, it may be an outlier or come from a new group about which there is currently no information. Conversely, if the classes are not well separated, some future observations may have small distances from more than one class. In such cases, it may again be undesirable to decide on a single possible class; instead two or more groups may be listed as possible *loci* for the observation.

The average improvement in motion compensation $\overline{IMC}_k(dB)$ between two consecutive frames, if $S$ is the frame being currently analyzed, is given by

$$\overline{IMC}_k(dB) = 10\log_{10}\left\{\frac{\sum_{\mathbf{r}\in\mathbf{S}}[I_k(\mathbf{r})-I_{k-1}(\mathbf{r})]^2}{\sum_{\mathbf{r}\in\mathbf{S}}[I_k(\mathbf{r})-I_{k-1}(\mathbf{r}-\mathbf{d}(\mathbf{r}))]^2}\right\}.$$

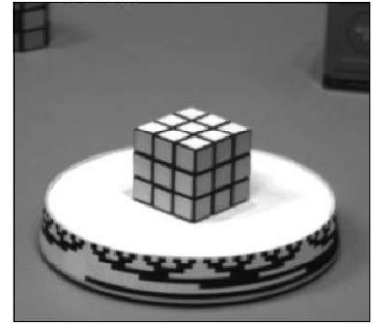

(a)

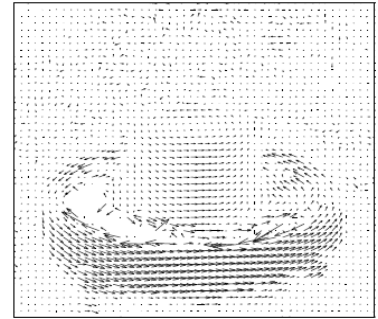

(b)

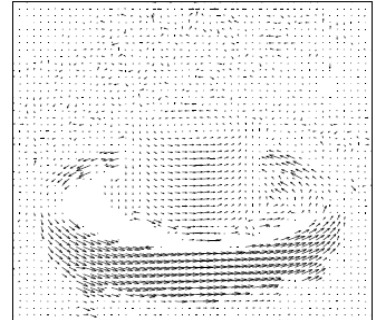

(c)

Fig. 5. Displacement field for the Rubik Cube sequence:. (a) Frame of Rubik Cube Sequence; (b) Corresponding displacement vector field for a 31×31 mask obtained by means of PCR$_1$ with SNR=20 dB; and (c) PCR$_2$, 31×31 mask with SNR=20 dB.

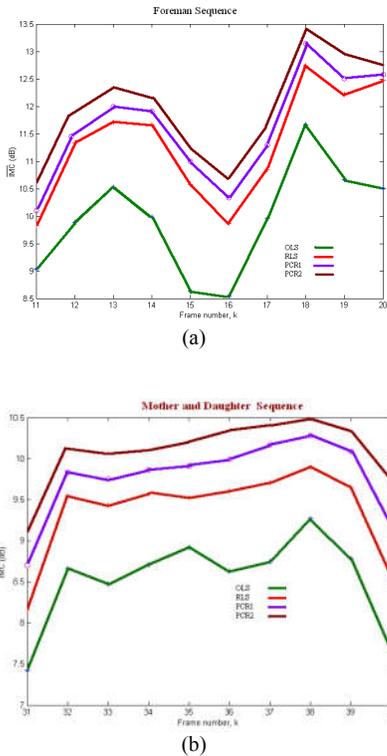

(a)

(b)

Fig. 4. Improvement in motion compensation curves for the "Foreman" and "Mother and Daughter" sequences.

At this time, partial analyses for the PCR$_1$ and the PCR$_2$ motion estimation procedures have been made for some video sequences. It is a known fact that there is a plethora of metrics to evaluate both quality as well as deviations from the original or expected DVFs. The ones used in this work are described below in conjunction with some results.

For a sequence of $K$ frames, the $\overline{IMC}(dB)$ becomes [4, 5]:

$$\overline{IMC}(dB) = 10\log_{10}\left\{\frac{\sum_{k=2}^{K}\sum_{\mathbf{r}\in\mathbf{S}}[I_k(\mathbf{r})-I_{k-1}(\mathbf{r})]^2}{\sum_{k=2}^{K}\sum_{\mathbf{r}\in\mathbf{S}}\left[I_k(\mathbf{r})-I_{k-1}\left(\mathbf{r}-\mathbf{d}(\mathbf{r})\right)\right]^2}\right\}.$$

When it comes to motion estimation, one seeks algorithms that have high values of $\overline{IMC}(dB)$. If the detected motion had no error, the denominator of the previous expression would be zero (perfect registration of motion) leading to $\overline{IMC}(dB) = \infty$.

Fig. 4 illustrates the evolution of $\overline{IMC}_k(dB)$ as a function of the frame number for two noiseless sequences: "Foreman" and "Mother and Daughter". These plots have been obtained without noise. PCR$_2$ works outperforms the other estimators due to the use of regularization in the PC domain

The SNR is defined as shown in [4, 5]:
$$SNR = 10\log_{10}(\sigma^2 / \sigma_n^2).$$

$\sigma^2$ and $\sigma_n^2$ stand for, respectively, the variances of the original image as well as the amount of noise corrupting a given frame.

Fig. 5 shows the DVFs for the "Rubik Cube" sequence with SNR=20 dB.

## V. CONCLUSION

In this paper, two PCR frameworks for the detection of motion fields are discussed. Both algorithms combine regression and PCA. The resulting transformed variables are uncorrelated. Unlike other works ([8, 11, 12]), we are not interested in reducing the dimensionality of the feature space describing different types of motion inside a neighbourhood surrounding a pixel. Instead, we use them in order to validate motion estimates. They can be seen as simple alternative ways of dealing with mixtures of motion displacement vectors. PCR$_1$ and PCR2 performed better than RLS estimators for noiseless and noisy images. More experiments are still needed in order to test the proposed algorithms with different types and levels of noise, so that the classification can be improved. It is also necessary to incorporate more statistical information in our models and to analyze if this knowledge will improve the outcome.

### ACKNOWLEDGEMENTS

The authors are thankful to CAPES, FAPERJ and CNPq for the financial support received.